\UseRawInputEncoding
\documentclass[pdflatex,sn-mathphys-num]{sn-jnl}

\usepackage[utf8]{inputenc}
\usepackage{bbm}
\usepackage{graphicx}%
\usepackage{multirow}%
\usepackage{amsmath,amssymb,amsfonts}%
\usepackage{amsthm}%
\usepackage{mathrsfs}%
\usepackage[title]{appendix}%
\usepackage{xcolor}%
\usepackage{textcomp}%
\usepackage{manyfoot}%
\usepackage{booktabs}%
\usepackage{algorithm}%
\usepackage{algorithmicx}%
\usepackage{algpseudocode}%
\usepackage{listings}%
\usepackage{amsmath, amssymb, amsfonts, mathtools, bm}
\usepackage{graphicx}
\usepackage{booktabs}
\usepackage{svg}
\usepackage[section]{placeins}

\newcommand{\R}{\mathbb{R}}

\newcommand{\concat}{\,\Vert\,}
\newcommand{\vv}[1]{\bm{#1}}
\newcommand{\mrm}[1]{\mathrm{#1}}

\theoremstyle{thmstyleone}%
\theoremstyle{thmstyletwo}%

\theoremstyle{thmstylethree}%

\begin{document}

\title[A Climate-Aware Deep Learning Framework for Generalizable Epidemic Forecasting]{A Climate-Aware Deep Learning Framework for Generalizable Epidemic Forecasting}


\author[1]{\fnm{Jinpyo} \sur{Hong}}\email{jinpyo\_hong@alumni.brown.edu}

\author[2,3]{\fnm{Rachel E.} \sur{Baker}}\email{rebaker@brown.edu}

\affil[1]{\orgdiv{School of Engineering}, \orgname{Brown University}, \orgaddress{\city{Providence}, \state{RI}, \country{USA}}}

\affil[2]{\orgdiv{Department of Epidemiology}, \orgname{School of Public Health}, \orgaddress{\city{Providence}, \state{RI}, \country{USA}}}

\affil*[3]{\orgdiv{Institute at Brown for Environment and Society}, \orgname{Brown University}, \orgaddress{\city{Providence}, \state{RI}, \country{USA}}}


\abstract{Precise outbreak forecasting of infectious diseases is essential for effective public health responses and epidemic control. The increased availability of machine learning (ML) methods for time-series forecasting presents an enticing avenue to enhance outbreak forecasting. Though the COVID-19 outbreak demonstrated the value of applying ML models to predict epidemic profiles, using ML models to forecast endemic diseases remains underexplored. In this work, we present ForecastNet-XCL (an ensemble model based on XGBoost+CNN+BiLSTM), a deep learning hybrid framework designed to addresses this gap by creating accurate multi-week RSV forecasts up to 100 weeks in advance based on climate and temporal data, without access to real-time surveillance on RSV. The framework combines high-resolution feature learning with long-range temporal dependency capturing mechanisms, bolstered by an autoregressive module trained on climate-controlled lagged relations. Stochastic inference returns probabilistic intervals to inform decision-making. Evaluated across 34 U.S. states, ForecastNet-XCL reliably outperformed statistical baselines, individual neural nets, and conventional ensemble methods in both within- and cross-state scenarios, sustaining accuracy over extended forecast horizons. Training on climatologically diverse datasets enhanced generalization furthermore, particularly in locations having irregular or biennial RSV patterns. ForecastNet-XCL's efficiency, performance, and uncertainty-aware design make it a deployable early-warning tool amid escalating climate pressures and constrained surveillance resources.}




\maketitle

\section{Main}\label{sec1}

Endemic respiratory diseases such as the respiratory syncytial virus (RSV) pose a persistent burden on global health systems, particularly among infants and the elderly. Unlike pandemic threats that emerge rapidly and globally, endemic pathogens display periodic, climate-dependent patterns locally defined by environmental, demographic, and infrastructural variables \cite{baker2020susceptible}. Although the COVID-19 pandemic spurred widespread advances in infectious disease modeling, including computational tools for near-term forecasting \cite{krymova2022trend} and policy impact \cite{kraemer2020effect} \cite{chinazzi2020effect}-the majority of this progress focused on pandemics. For endemic diseases, especially those modulated by climate, advances have lagged. Despite growing evidence that meteorological drivers strongly shape transmission, climate-informed endemic forecasting remains underdeveloped \cite{baker2022infectious}.

Climate conditions, including temperature, humidity, and precipitation, may affect virus persistence, host susceptibility, and transmission-driving behavioral patterns \cite{thongpan2020respiratory} \cite{moriyama2020seasonality}. For RSV in the United States, these drivers strongly influence epidemic timing and severity \cite{reis2016retrospective} \cite{pitzer2015environmental}. Standard statistical models - such as integrated autoregressive moving average (ARIMA), seasonal autoregressive models (SARIMA), and generalized linear models (GLM) - impose stationarity and linearity \cite{nobre2001dynamic}, restricting generalizability to years of climatic anomalies or disruption of behavior (e.g., during 2020-2021 NPI). Mechanistic models like Susceptible-Infected-Recovered (SIR) model transmission \cite{anderson1991infectious} based on fixed attributes and seasonal forcing \cite{grenfell2002dynamics}, but frequently embed seasonality as fixed sine or cosine terms and have limited mechanism to incorporate climate feedback or real-time exogenous data streams \cite{wagner2025societal}.

RSV offers a compelling test case for climate-informed machine learning (ML) forecasting. It is a well-characterized virus with clearly defined seasonal trends, high climate sensitivity, and significant regional heterogeneity \cite{baker2019epidemic}. For example, biennial epidemic cycles have been documented in northern states such as Minnesota, while southern regions such as Florida exhibit more regular annual outbreaks\cite{pitzer2015environmental}. These differences reflect underlying variation in climate, population density, mobility, and healthcare access \cite{ye2025understanding}. Although deep-learning models-such as LSTM \cite{chimmula2020time} \cite{wang2020time}, CNN \cite{pandianchery2023centralized} and LLM \cite{du2025advancing}-have advanced time-series forecasting primarily in COVID-19 contexts, most treat climate as peripheral lagged covariates without modeling mechanistic influence. For RSV specifically, prior deep-learning work has emphasized short-range horizons or onset prediction rather than long-horizon incidence trajectories \cite{yonekura2025prediction}. The key challenge is whether ML systems can generalize across spatial contexts and faithfully reproduce the complex, climate-modulated patterns observed in endemic transmission.

\begin{figure}[b!]
\centering
\includegraphics[width=0.9\textwidth]{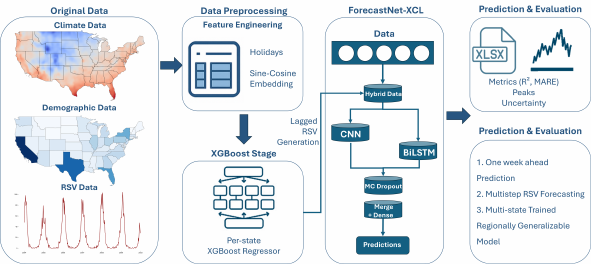}
\caption{\textbf{Schematic of ForecastNet-XCL.} From left to right: inputs combine weekly climate fields, state-level demographics, and RSV surveillance; preprocessing adds calendar features (for example, holidays) and seasonal embeddings. The architecture is two-stage: Stage 1 trains gradient-boosted trees (XGBoost) on recent covariates to predict next-week incidence, whose shifted predictions supply label-free autoregressive lags; Stage 2 is a hybrid CNN-BiLSTM that ingests covariates and generated lags over a 16-week window and produces multi-week trajectories via strictly recursive rollouts. Convolutions capture short-range temporal structure while recurrent units capture seasonal and inter-annual dependence; uncertainty is quantified with Monte Carlo dropout(see Methods for full details).}\label{fig1}
\end{figure}

In this study, we introduce ForecastNet-XCL, a unified, label-free, climate-aware framework for forecasting endemic respiratory diseases under operational constraints (Fig. 1). Rather than relying on future incidence, the method uses recent meteorological and calendar signals to generate calibrated probabilistic trajectories across heterogeneous regions. Using RSV across climatically diverse US states, we evaluate performance under three surveillance-aligned tasks: one-week-ahead accuracy, multi-step horizons (error growth and phase preservation), and cross-state generalization-under identical inputs and label-free training/testing to prevent leakage. The design emphasizes scalability and transferability, providing a template for climate-sensitive endemic respiratory forecasting.

\section{Results}\label{sec2}

\subsection{One-week-ahead Predictions}\label{subsec1}

\begin{figure}[h]
\centering
\includegraphics[width=0.9\textwidth]{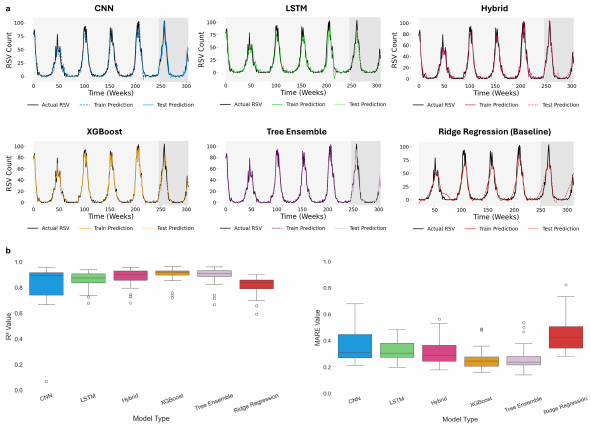}
\caption{\textbf{One-week-ahead RSV forecasting using state-specific training.}
\textbf{a}, Observed RSV incidence curves (solid black lines) in Arkansas overlaid with one-week-ahead forecasts (dashed lines) generated by six models: CNN, LSTM, hybrid CNN-LSTM, XGBoost, stacked tree ensemble, and ridge regression. The darker gray shaded region denotes the test period, while light gray shaded portions correspond to model fit on the training data. 
\textbf{b}, Boxplot quantitative comparison of forecasting accuracy using coefficient of determination ($R^2$, left) and Mean Absolute Relative Error (MARE, right) over 34 states.}
\label{fig2}
\end{figure}

In the first task, we assessed whether deep-learning and machine-learning models can forecast RSV incidence one week ahead using both climate covariates and recent RSV observations. Ground-truth inputs—meteorological variables and RSV incidence—were provided over a 16-week window, and models predicted the subsequent week’s incidence. This design approximates short-term surveillance settings in which near–real-time case reports are available. We evaluated model performance under both within-state and cross-state splits to assess generalizability, using data from 34 states with at least six consecutive years of surveillance observations.

We compared six approaches: two neural baselines (LSTM \cite{hochreiter1997long} and CNN \cite{lecun2015deep}), a hybrid CNN–LSTM, two tree-based ensembles (XGBoost \cite{chen2016xgboost} and a tree-based stacked ensemble \cite{galicia2019multi}), and a regularized linear baseline (ridge regression \cite{hoerl1970ridge}). All models used identical 16-week temporal input windows. Quantitative summaries appear in Figure~2b; representative forecasts for Arkansas are shown in Figure~2a.

Across models, XGBoost and the stacked ensemble achieved the highest accuracy. Under within-state training, XGBoost reached a mean $R^{2}$ of 0.91 (95\% CI, 0.89–0.93) with mean MARE 0.26 (95\% CI, 0.24–0.30), closely followed by the stacked ensemble (mean $R^{2}=0.89$, MARE $=0.27$). The hybrid CNN–LSTM also performed strongly (mean $R^{2}=0.88$, 95\% CI, 0.86–0.91; MARE $=0.31$) and exhibited a comparatively narrow interquartile range, indicating more consistent accuracy across states. Pure deep-learning models—CNN (mean $R^{2}=0.82$, MARE $=0.38$) and LSTM ($R^{2}=0.86$, MARE $=0.33$)—showed wider error distributions. Ridge regression was least accurate overall ($R^{2}=0.82$, MARE $=0.46$), reflecting limitations in capturing nonlinear dynamics.

Qualitatively (Fig.~2a), XGBoost and the stacked ensemble tracked seasonal peaks and troughs with high phase fidelity and amplitude precision. CNN and LSTM recovered broad shapes but occasionally misaligned peak onset or smoothed sharp surges. The hybrid CNN–LSTM tempered noise while retaining responsiveness to abrupt changes.

For cross-state generalization—each state held out entirely for testing—tree-based methods again led: XGBoost yielded the highest mean $R^{2}$ (0.88) and a low, stable MARE (0.32; 95\% CI, 0.31–0.32), followed by the stacked ensemble (mean $R^{2}=0.86$, MARE $=0.27$). Among neural models, the hybrid CNN–LSTM generalized best (mean $R^{2}=0.69$, MARE $=0.32$), outperforming LSTM ($R^{2}=0.67$, MARE $=0.51$) and CNN ($R^{2}=0.62$, MARE $=0.59$). These patterns suggest that convolutions capture short-range temporal structure while recurrent units encode longer-range dependencies, yielding more transferable features than either component alone.

\subsection{Multistep RSV Forecasting}\label{subsec2}

While tree-based learners such as XGBoost delivered strong one-week-ahead accuracy (Task 1), their performance degraded in recursive, multi-week prediction—conditions that mirror real‐world deployment. In this setting, models must iteratively generate incidence values without access to future ground-truth observations—a regime that exposes limits in temporal generalization, compounds error over time, and risks structural drift. Nonparametric trees are powerful at capturing nonlinearities in static inputs, but they are effectively memoryless; as a consequence, they can overfit local temporal idiosyncrasies and struggle to sustain trajectory fidelity over long horizons or during epidemiological regime shifts.

Motivated by our Task 1 finding that combining CNN and LSTM improved single-step accuracy, we designed ForecastNet-XCL for the harder recursive task by fusing a tree-based encoder with a deeper temporal network. ForecastNet-XCL comprises an XGBoost pre-module that learns nonlinear climate-to-incidence lag structure, followed by a CNN–BiLSTM backbone with self-attention. The CNN layers provide short-range sensitivity and denoising; the bidirectional LSTM supplies long-range temporal memory; attention reweights salient periods. Importantly, the model operates with only a 16-week look-back and never consumes future RSV labels at inference, reducing data requirements and improving deployment feasibility.

We evaluated all models under fully recursive inference across 34 U.S. states. Ground-truth incidence was used only within the initial input window, and subsequent steps relied exclusively on model-generated predictions. To establish a time-series-aware statistical baseline for recursive forecasting, we replaced ridge regression with a Seasonal ARIMA (SARIMA) model \cite{box1976analysis}. Unlike ridge, SARIMA explicitly captures autoregressive and seasonal dynamics, providing a closer representation of multi-step epidemiological signals and serving as a widely used benchmark in infectious disease and environmental forecasting.
\begin{figure}[h]
\centering
\includegraphics[width=0.9\textwidth]{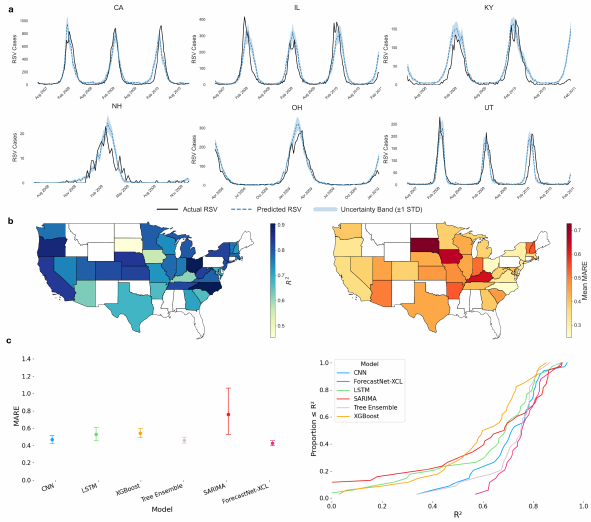}
\caption{\textbf{Recursive multi-step RSV forecasting performance across state-level scenarios.
}
\textbf{a}, Forecasted RSV incidence curves (dashed lines) generated by the recursive autoregressive ForecastNet-XCL, plotted against the observed RSV values (solid black line) in various states. Each panel corresponds to the held-out final 30 \% of each state’s time series, reflecting the heterogeneous reporting durations across 34 states. Shaded regions denote 95\% Monte Carlo dropout-based uncertainty intervals. 
\textbf{b}, Distribution of $R^2$ (left) and MARE (right) for ForecastNet-XCL across all 34 states under recursive forecasting conditions.
\textbf{c}, Comparative performance metrics for all models in the recursive setting, shown as metric range plots and cumulative distribution functions.
}
\label{fig3}
\end{figure}
ForecastNet-XCL accurately recovered peak timing, peak magnitude, and post-peak decay over 52-week horizons (Fig.~\ref{fig3}a), without cumulative drift or phase distortion. States such as Arkansas and Pennsylvania illustrate tight alignment of peak onsets and trough recoveries—even in seasons with irregular behavior.

Quantitatively, ForecastNet-XCL achieved the best overall accuracy, leading on both $R^2$ and MARE distributions (Fig.~\ref{fig3}b–c). Among baselines, stacked tree ensembles were the strongest competitors in the recursive regime, narrowly outperforming standalone deep nets on median performance but exhibiting greater variance and more frequent peak-timing lag. Models lacking sufficient temporal depth tended to smooth sharp seasonal inflections or react with delay, consistent with error accumulation in iterative forecasting.

ForecastNet-XCL's performance remained geographically consistent. Accuracy declined in low-incidence, weak-seasonality states (e.g., Vermont), where high weekly volatility reduces signal-to-noise, yet ForecastNet-XCL preserved coherent seasonal shape and avoided divergence. This stability reflects the complementary design: the XGBoost pre-module extracts nonlinear climate–lag structure, while the CNN–BiLSTM with attention maintains temporal continuity, reducing overshoot and phase lag common in purely deep or purely tree-based recursive models.

Together, these results support ForecastNet-XCL as a practical engine for real-time pipelines: it produces multi-week forecasts from recent climate and calendar inputs alone—without future RSV labels—scales to long autoregressive horizons, and generalizes across diverse climates. The empirical ranking in Fig.~\ref{fig3}c further clarifies our design choice: after observing in Task 1 that hybrid CNN–LSTM architectures improved single-step accuracy, we extended the idea by coupling a strong tree-based encoder with a deeper CNN–BiLSTM forecaster for recursive inference, yielding state-of-the-art performance with robust temporal stability.

\subsection{Multi-state Transfer Learning}\label{subsec3}
\begin{figure}[h]
\centering
\includegraphics[width=0.9\textwidth]{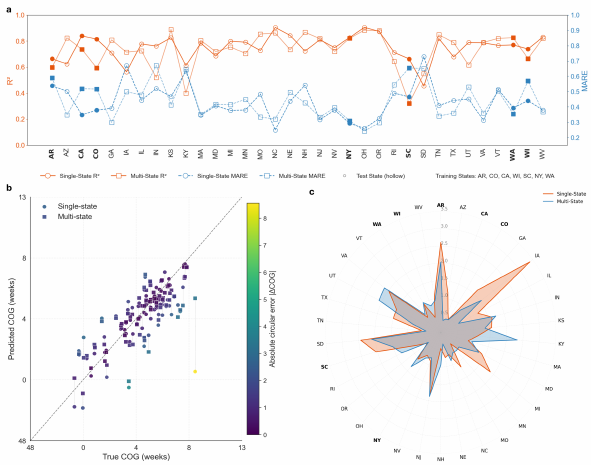}
\caption{\textbf{Recursive multi-step RSV forecasting performance across single-state and multi-state trained scenarios.
}
\textbf{a}, State-by-state comparison of forecasting accuracy, with coefficients of determination ($R^2$, left axis, orange) and mean absolute relative errors (MARE, right axis, blue) plotted for single-state (solid markers) and multi-state (hollow markers) training.
\textbf{b}, Predicted versus observed center-of-gravity (COG) of the RSV season for each state–season under single-state (circles) and multi-state (squares) training. Points are colored by absolute circular timing error $|\Delta\mathrm{COG}|$ (weeks).
\textbf{c}, Per-state timing error shown on a radial axis for single-state (orange) and multi-state (blue) models; values nearer 0 indicate better phase alignment. Bold labels mark the seven pretraining states (AR, CA, CO, SC, NY, WA, WI).
}
\label{fig4}
\end{figure}
To test how training-data diversity shapes generalization, we compared two ForecastNet-XCL configurations. In the \emph{single-state} setting, a separate model was trained and evaluated on each state’s series, approximating idealized local conditions with ample surveillance but no geographic exposure. In the \emph{multi-state} setting, we trained a base model on pooled data from seven climatically and demographically diverse states (AR, CO, CA, WI, SC, NY, WA), using state embeddings to capture both shared structure and state-specific idiosyncrasies. The pooled model was then (i) applied directly to the seven training states and (ii) fine-tuned on each remaining state’s local training split before testing. This design probes whether exposure to heterogeneous outbreak dynamics, followed by lightweight local adaptation, improves accuracy and robustness in data-limited or climatically distinct regions.

By conventional summary metrics ($R^2$, MARE), the two configurations perform similarly in most settings (Fig.~\ref{fig4}a): both recover the dominant seasonal signature of RSV with overlapping error distributions. Apparent parity at this aggregate level, however, masks salient differences in stability, robustness, and epidemiological fidelity.
\begin{figure*}[t]
\centering
\includegraphics[width=0.95\textwidth]{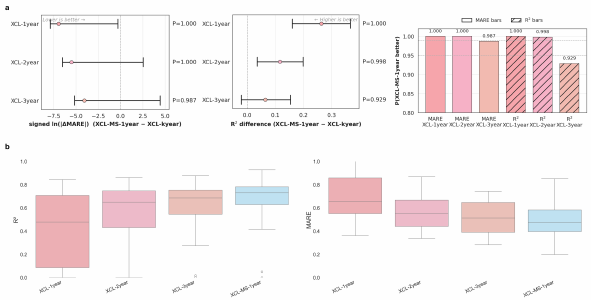}
\caption{\textbf{Multi-state transfer learning (XCL) improves accuracy and signal quality.}
\textbf{a}, Summary of performance gains. \emph{Left}, MARE signed–log difference
$\mathrm{sgn}(\Delta)\ln|\Delta\mathrm{MARE}|$ with
$\Delta\mathrm{MARE}=\mathrm{MARE}_{\text{XCL-MS-1year}}-\mathrm{MARE}_{\text{XCL-}k\text{year}}$,
$k\in\{1,2,3\}$ (points: bootstrap means over 10{,}000 iterations; lines: 95\% CIs; right labels: $P(\text{XCL-MS-1year better})$).
\emph{Middle}, $R^2$ difference with the same bootstrap summary.
\emph{Right}, bootstrap win probabilities $P(\text{XCL-MS-1year better})$ for each metric and horizon
\textbf{b}, \emph{Left}, MARE boxplots; \emph{right}, $R^2$ boxplots for
ForecastNet-XCL-1year, ForecastNet-XCL-2year,ForecastNet-XCL-3year, and ForecastNet-XCL-MS-1year 
\textbf{Notation:}, For brevity in panel labels, \textit{ForecastNet-XCL} (single-state) and \textit{ForecastNet-XCL-MS} (multi-state) are abbreviated as \(\text{XCL-}k\text{year}\) and \(\text{XCL-MS-1year}\), respectively.} 
\label{fig5}
\end{figure*}

The multi-state model shows a decisive advantage in timing accuracy. As quantified in Fig.~\ref{fig4}b–c using the center of gravity (COG) of seasonal peaks, pooled training yields lower circular error than the single-state baseline. Intuitively, exposure to diverse temporal signatures regularizes the model against overfitting to local anomalies, improving phase alignment. Timing precision is operationally critical: while amplitude errors chiefly affect burden estimates, timing errors misalign vaccination campaigns, prophylaxis windows, and hospital capacity planning. Thus, improved timing fidelity represents a substantive epidemiological gain even when $R^2$ and MARE appear comparable.

To evaluate model robustness under data scarcity and temporal non-stationarity, we stress-tested both training regimes. The models were trained on only the first one, two, or three years of each state’s time series and evaluated exclusively on the final 30\% of held-out data. This design imposes two stringent constraints: severe data limitation from a short local history and a multi-year gap between the training and testing periods, which forces the model to generalize temporally rather than rely on seasonal persistence. As several states have around six years of total surveillance data, the three-year window represents the maximum feasible training history for the single-state models.

Under these challenging conditions, the multi-state model demonstrated a decisive performance advantage. A multi-state model fine-tuned on only one year of local data (\textit{ForecastNet-XCL-MS-1year}) consistently outperformed single-state baselines trained on one, two, or even three years of data (\textit{ForecastNet-XCL-}$k$\textit{year}, where $k \in \{1,2,3\}$). Bootstrap analyses show that the signed–log difference in Mean Absolute Relative Error (MARE) is uniformly negative, while the difference in $R^2$ is uniformly positive across all training horizons (Fig.~\ref{fig5}a). The corresponding 95\% confidence intervals robustly exclude zero, and the probabilities of the multi-state model being superior approach 100\%. Furthermore, the across-state performance distributions confirm this trend, revealing a clear shift toward lower MARE and higher $R^2$ with comparable or reduced dispersion, indicating improved stability and central tendency (Fig.~\ref{fig5}b).

These findings demonstrate that transfer learning from a geographically and climatically diverse dataset, combined with brief local fine-tuning, can effectively substitute for longer local training histories. The ability of a ForecastNet-XCL-MS trained on 1 year of data to match or exceed the performance of a three-year trained Forecast-XCL, despite a pronounced temporal discontinuity, underscores its resilience. This property is critical for developing and deploying reliable forecasting systems in jurisdictions with sparse or interrupted surveillance records.

\section{Discussion}\label{sec3}

This study presents ForecastNet-XCL, a climate-aware, strictly recursive forecasting framework intended to address a practical gap in endemic respiratory disease modeling: producing multi-week incidence trajectories when future case data are unavailable. The approach combines a tree-based module that synthesizes label-free autoregressive signals from recent meteorology with a convolutional–recurrent backbone that encodes short- and longer-range temporal structure. Relative to conventional statistical models (e.g., ARIMA/SARIMA) and compartmental frameworks (e.g., SIR/SIRS), which often assume stationarity or depend on contemporaneous observations \cite{pan2016arima,satrio2021time,toda2020susceptible,cooper2020sir,atkeson2020estimating}, ForecastNet-XCL aims to learn representations from exogenous drivers alone, aligning more closely with scenarios in which epidemiological reporting is delayed or intermittent.

A deliberate design choice is parsimony in historical context: ForecastNet-XCL uses a 16-week lookback window yet, in our experiments, generated coherent longer-horizon forecasts. This compact input reduces data and engineering burden and mitigate label leakage risks during evaluation. Despite the truncated context, the model performed competitively under fully recursive rollouts, tending to preserve seasonal timing, peak magnitude, and post-peak decay in most states. Performance was weaker in low-signal environments (e.g., limited seasonality or low incidence), where fine-grained tracking is challenging for any model; nonetheless, ForecastNet-XCL generally avoided error drift toward implausible trajectories, which is encouraging for prospective use.

Evidence from pooled multi-state training suggests the model can learn patterns that transfer across locations. Training on climatically diverse states with state embeddings did not degrade within-state results and was associated with improved peak-timing estimates when transferred or lightly fine-tuned. In particular, the multi-state configuration reduced circular errors in the seasonal center of gravity, a practically meaningful improvement because timing affects advance procurement, prophylaxis scheduling, and capacity planning more directly than small gains in pointwise error. Stress tests with short and fragmented training histories further indicated that exposure to heterogeneous outbreaks can stabilize learning where single-state models became variable. Taken together, these findings support (but do not prove) a useful design principle under data limitations: leverage cross-context diversity to regularize temporal representations, then adapt modestly to local conditions.

Uncertainty quantification is an essential component for decision support. We used Monte Carlo dropout as a lightweight Bayesian approximation \cite{gal2016dropout} and observed empirically reasonable calibration across climates and seasons. For higher-stakes deployments, deeper calibration could be explored without altering the overall architecture—for example, deep ensembles \cite{lakshminarayanan2017simple}, post-hoc trajectory-level calibration (e.g., isotonic or Platt scaling), or hierarchical variance pooling to share information across neighboring regions. Complementing $R^2$ and MARE with proper scoring rules (e.g., interval scores or CRPS) would also provide a more complete assessment of reliability when forecasts feed threshold-based policies.

Several limitations temper our conclusions. First, the retrospective evaluation used observed meteorology at test time; operational pipelines must substitute forecasted fields. This replacement is feasible for key drivers (temperature and precipitation) given routine availability from systems such as GFS \cite{noaa-gfs-products}, ECMWF \cite{ecmwf-forecasts}, and CPC \cite{cpc-monthly-seasonal}, but the impact of meteorological forecast error on epidemic predictions remains to be quantified. Future work should propagate weather-forecast uncertainty via multi-scenario forcing or training against ensembles of meteorological predictions. Second, although we evaluated across 34 U.S. states, generalization beyond this setting (other countries, sub-state geographies, or diseases) remains an open question. Scaling to finer spatial units will require handling sparsity and local nonstationarities; potential avenues include graph-aware convolutions, hierarchical training, or spatiotemporal weight sharing. Third, while we observed benefits from pooled training, more systematic ablations (e.g., removing the tree-based lag generator, varying the lookback window, or swapping recurrent components) would clarify which design elements are most responsible for stability.

Although this work focuses on RSV, the ingredients of ForecastNet-XCL—synthetic autoregressive memory from exogenous drivers, a compact temporal receptive field, and state-aware transfer—are disease-agnostic and could be adapted to influenza, enteroviruses, or other climate-sensitive pathogens, provided exogenous signals with plausible mechanistic links are available and strict anti-leakage protocols are maintained. Overall, our results suggest that climate-aware, label-free architectures can be viable components of early-warning systems when real-time case data are delayed. We view ForecastNet-XCL as a step in that direction, with immediate priorities including prospective evaluation with forecasted meteorology, expanded external validation, and stronger uncertainty calibration to support operational decision-making \cite{bodin2017latent}.

\section{Methods}\label{sec4}
\subsection{Dataset Construction and Preprocessing}\label{subsec4}
The research utilizes a multi-source, state-based dataset that combines weekly respiratory syncytial virus (RSV) hospitalization data with climate and demographic factors within 42 states of the US. Every record is associated with a single epidemiological week, allowing rigorous analysis of temporal disease patterns under diverse environmental and population contexts.

\subsection{Epidemiological Data}\label{subsec5}
The core outcome variable—weekly RSV incidence—was extracted from the State Inpatient Databases (SIDs), curated under the Healthcare Cost and Utilization Project (HCUP) by the U.S. Agency for Healthcare Research and Quality (AHRQ). These records provide comprehensive weekly aggregates of RSV-related hospitalizations across participating states, offering standardized temporal resolution and sufficient granularity to model intra- and inter-annual variation in virus transmission \cite{abu2023has}.

\subsection{Climate Variables}\label{subsec6}
Environmental drivers of RSV transmission were incorporated using historical station-level weather data from the National Oceanic and Atmospheric Administration (NOAA) Climate Data Online archive. Daily values for average, maximum, and minimum temperature (TOBS, TMAX, TMIN), precipitation (PRCP), snowfall (SNOW), snow depth (SNWD), and wind speed (AWND) were aggregated to the weekly level and averaged across all meteorological stations within each state. These weekly aggregates allowed environmental variation to be consistently aligned with RSV epidemiological records while preserving climatic diversity between states.

\subsection{Data Alignment and Feature Engineering}\label{subsec7}
We engineer domain-specific features from raw meteorological and calendar data. Calendar features include a binary U.S. holiday indicator $h_t$ and cyclic week-of-year encoding:
\[
    s_t^{\sin} = \sin(2\pi w_t/52), \quad s_t^{\cos} = \cos(2\pi w_t/52),
\]
where $w_t \in \{1,\ldots,52\}$ is the week number. Epidemiological features capture disease-relevant conditions:
\begin{align}
    \text{extreme\_cold}_t &= \mathbb{1}[\text{TMIN}_t < q_{10}], \\
    \text{temp\_range}_t &= \text{TMAX}_t - \text{TMIN}_t, \\
    \text{precip\_intensity}_t &= \text{PRCP}_t \times \text{AWND}_t,
\end{align}
where $q_{10}$ is the 10th percentile of minimum temperature. Meteorological variables $\{\text{TMIN}, \text{TMAX}, \text{PRCP}\}$ are lagged by $\{7, 14\}$ weeks to capture delayed environmental effects.

To prevent target leakage, we generate synthetic RSV lags using Stage-1 XGBoost predictions. For test indices $t \geq t_{\text{test}}^{\min}$, the synthetic lags are:
\[
    \tilde{y}_t^{(\ell)} = \hat{y}^{\text{xgb}}_{t-\ell}, \quad \ell \in \{1,2,3,4\},
\]
while training indices use actual lagged values $y_{t-\ell}$. The complete feature vector at time $t$ is:
\[
    \tilde{\mathbf{x}}_t = [\text{base meteorology}, \text{calendar}, \text{epidemiological}, \text{weather lags}, \tilde{y}_t^{(1:4)}] \in \mathbb{R}^d,
\]
with $d \approx 24$ depending on feature availability per state.

All features undergo MinMax scaling using training statistics: $\tilde{f}_t = (f_t - f_{\min}^{\text{train}})/(f_{\max}^{\text{train}} - f_{\min}^{\text{train}})$. 

\subsection{ForecastNet-XCL Model Architecture and Recursive Forecasting Strategy}\label{subsec8}
Let $y_t\in\R$ denote weekly RSV incidence and $\vv{x}_t\in\R^{p}$ the exogenous feature vector (precipitation, temperature, snow depth, wind speed, county population, holiday indicator, and week-of-year sine/cosine encodings, optionally augmented with engineered interactions such as temperature range and precipitation–wind terms). With a four-week look-back we define the context
\[
  \vv{Z}_t=\bigl[\vv{x}_{t-3},\,\vv{x}_{t-2},\,\vv{x}_{t-1},\,\vv{x}_{t}\bigr]\in\R^{4p}.
\]
An XGBoost regressor $f_{\mrm{xgb}}(\cdot;\phi)$ is fit by
\begin{equation}
  \min_{\phi}\;\sum_{t\in\mathcal{T}_{\mrm{train}}}\Bigl(y_{t+1}-f_{\mrm{xgb}}(\vv{Z}_t;\phi)\Bigr)^2,
  \qquad
  \hat{y}^{\mrm{xgb}}_{t+1\mid t}=f_{\mrm{xgb}}(\vv{Z}_t;\phi).
\end{equation}
At evaluation time for this XGBoost, we compute, using only exogenous inputs, a quartet of \emph{synthetic incidence lags}
\begin{equation}
  \hat{y}^{(\ell)}_{t}=\hat{y}^{\mrm{xgb}}_{t-\ell},\qquad \ell\in\{1,2,3,4\},
\end{equation}
which emulate auto-regressive memory without referencing future labels and thus avoid leakage. Selected weather variables may also be lagged (e.g., 7- and 14-week shifts).

Prediction of the hybrid part is performed on a rolling 16-week window that concatenates exogenous features with the synthetic lags. For each $t$ we form
\begin{equation}
  \vv{H}_t=\bigl[\vv{x}_{t-15},\ldots,\vv{x}_{t}\concat \hat{y}^{(1)}_{t},\ldots,\hat{y}^{(4)}_{t}\bigr]\in\R^{16\times d},
\end{equation}
where $d$ is the per-week feature dimension after concatenation.
A convolutional pathway extracts localized temporal motifs with three parallel 1-D convolutions at kernel sizes $k\in\{2,4,8\}$ (128 then 64 filters; ReLU):
\begin{equation}
  \vv{U}_k=\mrm{Conv}^{(2)}_{k}\!\bigl(\mrm{Conv}^{(1)}_{k}(\vv{H}_t)\bigr),\qquad
  \vv{h}_{\mrm{cnn}}=\mrm{vec}\!\bigl(\vv{U}_2\concat\vv{U}_4\concat\vv{U}_8\bigr).
\end{equation}
In parallel, a bidirectional LSTM produces hidden states $\vv{S}\in\R^{16\times m}$ that are refined by multi-head self-attention (four heads) with a residual connection:
\begin{equation}
  \vv{A}=\mrm{MHA}(\vv{S},\vv{S},\vv{S}),\qquad \vv{S}'=\vv{S}+\vv{A},
\end{equation}
followed by a unidirectional LSTM to yield $\vv{h}_{\mrm{rnn}}\in\R^{32}$.
The fused representation is passed through dense layers with a skip connection to produce $hat{y}_{t+1}$.

\textbf{Optimization and validation.}
Training uses Adam (initial learning rate $\approx6\times10^{-4}$), cosine-annealed scheduling, and gradient clipping (clip-norm $=1.0$).
Each state uses a temporal 70/30 train/test split; within the $70\%$ training portion the last $20\%$ is held out for early stopping, and we perform 3-fold time-series cross-validation on training data only (For ForecastNet-XCL). Test sets remain untouched until final evaluation.
The forecasting protocol is recursive at inference: synthetic lags are precomputed from the exogenous regressor and the hybrid network never feeds back its own predictions, avoiding error drift while preserving auto-regressive memory.

\textbf{ForecastNet-XCL-MS with state embeddings.}
To enable cross-jurisdiction generalization, we learn an embedding matrix
\(\mathbf E\in\mathbb{R}^{S\times d_e}\) (with \(d_e=16\)) indexed by a state-ID map
\(m:\{1,\dots,S\}\!\to\!\{0,\dots,S\!-\!1\}\).
For a state \(s\), we retrieve the embedding \(\mathbf e_s=\mathbf E_{m(s):}\) and
repeat it along the temporal axis to match the sequence length \(L=16\):
\(\tilde{\mathbf E}_s=\mathrm{Repeat}(\mathbf e_s, L)\).
Given the per-step feature sequence \(\mathbf H_t\in\mathbb{R}^{L\times d}\),
the model consumes
\[
  \tilde{\mathbf H}^{(s)}_t \;=\; \bigl[\,\mathbf H_t \;\Vert\; \tilde{\mathbf E}_s\,\bigr],
  \qquad
  \hat y^{(s)}_{t+1} \;=\; g_{\theta}\!\bigl(\tilde{\mathbf H}^{(s)}_t,\;\mathbf e_s\bigr),
\]
where we also pass the \emph{static} \(\mathbf e_s\) forward via a head skip connection:
after a multi-scale CNN pathway and a BiLSTM\(+\)self-attention pathway produce
\(\mathbf h_{\mathrm{cnn}}\) and \(\mathbf h_{\mathrm{rnn}}\), the fused representation is
\[
  \mathbf z \;=\; \bigl[\,\mathbf h_{\mathrm{cnn}} \;\Vert\; \mathbf h_{\mathrm{rnn}} \;\Vert\; \mathbf e_s\,\bigr].
\]
Parameters \((\theta,\mathbf E)\) are pretrained jointly across source states by
\[
  \min_{\theta,\mathbf E}
  \sum_{s=1}^{S}\;\sum_{t\in\mathcal{T}^{(s)}_{\mathrm{train}}}
  \mathcal L\!\left(y^{(s)}_{t+1},\, g_{\theta}\!\bigl(\tilde{\mathbf H}^{(s)}_t,\mathbf e_s\bigr)\right),
\]
using Adam (base learning rate \(\approx 6\times 10^{-4}\)), early stopping on a temporal
validation split, and gradient clipping.

For a previously unseen state \(s^\ast\), we fine-tune the pretrained network at a reduced
rate \(\eta_{\mathrm{ft}}=10^{-4}\). Approximately \(70\%\) of layers are frozen while
keeping the \texttt{state\_embedding} layer \emph{trainable}:
\[
  \min_{\theta_{\mathrm{free}}}
  \sum_{t\in\mathcal{T}^{(s^\ast)}_{\mathrm{train}}}
  \mathcal L\!\left(y^{(s^\ast)}_{t+1},\,
  g_{\theta_{\mathrm{frozen}},\theta_{\mathrm{free}}}\!\bigl(\tilde{\mathbf H}^{(s^\ast)}_t,\mathbf e_{s^\ast}\bigr)
  \right).
\]

During model testing, we assume access to future climate variables over the forecast horizon. This design reflects the intended deployment context for ForecastNet-XCL as a climate-informed early warning tool—one that leverages meteorological forecasts, rather than real-time case data, to anticipate epidemic trends. While ground-truth climate values are used for offline evaluation, the model is intended to operate alongside existing environmental forecasting systems such as NOAA’s Global Forecast System (GFS) or ECMWF, which routinely provide short-, medium-, and even long-range climate forecasts (e.g., CPC seasonal outlooks). These systems offer reliable predictions for core variables such as temperature, precipitation, and snow depth with lead times of up to several weeks or months. We emphasize that the model does not require access to real-time RSV incidence at any point during inference and relies solely on climate and temporal inputs that are operationally feasible under real-world constraints.

\subsection{Uncertainty Quantification with Monte Carlo Dropout}\label{subsec9}

We estimate epistemic uncertainty using Monte Carlo Dropout. Dropout is kept active during both 
training and inference (by invoking dropout layers with \texttt{training{=}true}). At test time we perform 
$T = 50$ stochastic forward passes per timestep, each with an independently sampled dropout mask:
\begin{equation}
\hat{y}_t^{(i)} = f_{\text{hybrid}}\bigl(\mathbf{z}_t; \theta, m^{(i)}\bigr), \quad i = 1, \ldots, T,
\end{equation}
where $\mathbf{z}_t$ is the input sequence at time $t$, $\theta$ are the learned weights, and $m^{(i)}$ denotes the $i$-th dropout mask.

\medskip

For each timestep we summarize the predictive distribution by the sample mean and (population) standard deviation across the $T$ passes:
\begin{equation}
\bar{y}_t = \frac{1}{T} \sum_{i=1}^T \hat{y}_t^{(i)}, 
\qquad 
\sigma_t = \sqrt{\frac{1}{T} \sum_{i=1}^T \bigl(\hat{y}_t^{(i)} - \bar{y}_t \bigr)^2}.
\end{equation}
\medskip

We report empirical $95\%$ prediction intervals using the percentiles of the Monte Carlo samples:
\begin{equation}
\text{CI}_{95\%}^{\text{emp}}(t) = 
\left[
\operatorname{Quantile}_{2.5\%}\!\left(\left\{\hat{y}_t^{(i)}\right\}_{i=1}^T\right),
\;
\operatorname{Quantile}_{97.5\%}\!\left(\left\{\hat{y}_t^{(i)}\right\}_{i=1}^T\right)
\right].
\end{equation}
\medskip

All statistics are computed in the scaled space and then inverse-transformed to the original RSV scale for reporting and figures. Dropout rates follow the architecture: $0.2$ in CNN/dense branches and $0.3$ after the LSTM.

\subsection{Evaluation Metrics and Loss Metrics}\label{subsec10}
\subsubsection{Evaluation Metrics}

Let $\{y_t\}_{t=1}^N$ denote the ground-truth RSV counts and $\{\hat{y}_t\}_{t=1}^N$ the corresponding
predictions, all on the original (inverse--transformed) scale.

\paragraph{Mean Squared Error (MSE).}
\begin{equation}
\text{MSE} \;=\; \frac{1}{N} \sum_{t=1}^N (y_t - \hat{y}_t)^2.
\end{equation}

\paragraph{Coefficient of Determination ($R^2$).}
\begin{equation}
R^2 \;=\; 1 - \frac{\sum_{t=1}^N (y_t - \hat{y}_t)^2}{\sum_{t=1}^N (y_t - \bar{y})^2},
\qquad
\bar{y} \;=\; \frac{1}{N}\sum_{t=1}^N y_t.
\end{equation}

\paragraph{Mean Absolute Relative Error (MARE).}
\begin{equation}
\text{MARE} \;=\; \frac{\sum_{t=1}^N \lvert y_t - \hat{y}_t\rvert}{\sum_{t=1}^N y_t + \varepsilon},
\qquad \varepsilon = 10^{-8}.
\end{equation}

\noindent
Unlike the per--time point mean of ratios $\tfrac{1}{N}\sum_t \tfrac{|y_t-\hat{y}_t|}{y_t+\varepsilon}$, 
the above global form (ratio of sums) matches the implementation and is numerically stable when $y_t$  approaches zero during the off--season. All metrics are computed after fitting scalers on the training portion only and then inverse--transforming predictions to the original RSV scale. Chronological splits prevent information leakage; additionally, for the single--state pipeline we use forward--chaining time--series cross--validation on the training set.

\section{Data availability}\label{sec5}
RSV hospitalization data come from the State Inpatient Databases (SIDs) of the Healthcare Cost and Utilization Project (HCUP) maintained by the Agency for Healthcare Research and Quality (AHRQ). This data is available to researchers after signing a data use agreement. For access information, visit: https://hcup-us.ahrq.gov/sidoverview.jsp

Climate data are publicly available from NOAA's Global Historical Climatology Network (GHCN-Daily) at: https://www.ncei.noaa.gov/products/land-based-station/global-historical-climatology-network-daily

\section{Code availability}\label{sec6}
The source code is freely available via GitHub at: https://github.com/jinpyohong-blip/ForecastNet-XCL

\begin{appendices}




\end{appendices}

\FloatBarrier
\bibliography{sn-bibliography}

\end{document}